\documentclass[11pt]{article}

\usepackage{ACL2023}
\usepackage{times}
\usepackage{latexsym}

\usepackage[T1]{fontenc}

\usepackage[utf8]{inputenc}

\usepackage{microtype}

\usepackage{inconsolata}

\title{A dynamic programming algorithm \\ for span-based nested named-entity recognition in $\mathcal O(n^2)$}

\author{Caio Corro \\
  Universite Paris-Saclay, CNRS, LISN, 91400, Orsay, France \\
  \texttt{caio.corro@lisn.upsaclay.fr}}

\usepackage{verbatim}
\usepackage{amsmath}
\usepackage{amsfonts}
\usepackage{amssymb}
\usepackage{graphicx}
\usepackage{amsthm}
\usepackage{algpseudocode}
\usepackage{algorithmicx}
\usepackage{float}
\usepackage{todonotes}
\newfloat{algorithm}{t}{lop}
\usepackage{bussproofs}

\usepackage{booktabs}
\usepackage{verbatim}
\usepackage{stmaryrd}
\usepackage{caption}

\usepackage{tikz}
\usetikzlibrary{math,calc,positioning, fit,quotes,arrows.meta,patterns,graphs}
\usepackage{pgfplots}
\pgfplotsset{scaled y ticks=false}

\tikzset{>=latex}

\usepackage{amsmath,amsfonts,bm}

\def\eqref#1{equation~\ref{#1}}

\def\1{\bm{1}}

\def\vs{{\bm{s}}}

\def\vw{{\bm{w}}}

\def\vy{{\bm{y}}}

\def\evs{{s}}

\def\evw{{w}}

\def\evy{{y}}

\DeclareMathAlphabet{\mathsfit}{\encodingdefault}{\sfdefault}{m}{sl}
\SetMathAlphabet{\mathsfit}{bold}{\encodingdefault}{\sfdefault}{bx}{n}

\newcommand{\R}{\mathbb{R}}

\DeclareMathOperator*{\argmax}{arg\,max}

\begin{document}

\maketitle

\begin{abstract}
Span-based nested named-entity recognition (NER) has a cubic-time complexity using a variant of the CYK algorithm.
We show that by adding a supplementary structural constraint on the search space, nested NER has a quadratic-time complexity, that is the same asymptotic complexity than the non-nested case.
The proposed algorithm covers a large part of three standard English benchmarks and delivers comparable experimental results.
\end{abstract}

\section{Introduction}
\begin{figure*}[t]
    \begin{minipage}[t]{0.5\textwidth}
        \begin{tikzpicture}[
    every node/.style={
        rectangle,
        text height=1.5ex,
        text depth=.25ex
    }
]
    \node (he) [rectangle] { He};
    \node [circle, above=0.1cm of he, draw=black, inner sep=2pt] {1};
    
    \node (lost) [rectangle, right=0cm of he] {\texttt{lost}};
    \node (an) [rectangle, right=0cm of lost] {\texttt{an}};
    \node (election) [rectangle, right=0cm of an] {\texttt{election}};
    \node (to) [rectangle, right=0cm of election] {\texttt{to}};
    \node (a) [rectangle, right=0cm of to] {\texttt{a}};
    \node (dead) [rectangle, right=0cm of a] {\texttt{dead}};
    \node (man) [rectangle, right=0cm of dead] {\texttt{man.}};
    
    \draw(he.south west)--node[yshift=-0.25cm]{\textsc{Per}} (he.south east);
    \draw(a.south west)--node[yshift=-0.25cm]{\textsc{Per}}(man.south east);

    \draw[white ]([yshift=-0.6cm]he.south west)--node[yshift=-0.25cm]{ \textsc{Per}} ([yshift=-0.6cm]he.south east);
\end{tikzpicture}
    \end{minipage}%
    \begin{minipage}[t]{0.5\textwidth}
        \begin{tikzpicture}[
    every node/.style={
        rectangle,
        text height=1.5ex,
        text depth=.25ex
    }
]
    \node (this) [rectangle] {\texttt{This}};
    \node [circle, above=0.1cm of this, draw=black, inner sep=2pt] {2};
    \node (is) [rectangle, right=0cm of this] {\texttt{is}};
    \node (your) [rectangle, right=0cm of is] {\texttt{your}};
    \node (second) [rectangle, right=0cm of your] {\texttt{second}};
    \node (one) [rectangle, right=0cm of second] {\texttt{one}};
    \node (he) [rectangle, right=0cm of one] {\texttt{he}};
    \node (has) [rectangle, right=0cm of he] {\texttt{has}};
    \node (missed) [rectangle, right=0cm of has] {\texttt{missed.}};
    
    \draw(this.south west)--node[yshift=-0.25cm]{ \textsc{Per}} (this.south east);
    \draw([yshift=-0.6cm]your.south west)--node[yshift=-0.25cm]{ \textsc{Per}} ([yshift=-0.6cm]missed.south east);
    \draw (your.south west)--node[yshift=-0.25cm]{ \textsc{Per}} (your.south east);
    \draw (he.south west)--node[yshift=-0.25cm]{ \textsc{Per}} (he.south east);
\end{tikzpicture}
    \end{minipage}
    \vspace{-0.7cm}\begin{minipage}[t]{0.57\textwidth}
        \vspace{-0.5cm}\begin{tikzpicture}[
    every node/.style={
        rectangle,
        text height=1.5ex,
        text depth=.25ex
    }
]
    \node (he) [rectangle] {\texttt{He}};
    \node [circle, above=0.1cm of he, draw=black, inner sep=2pt] {3};
    \node (is) [rectangle, right=0cm of he] {\texttt{is}};
    \node (retired) [rectangle, right=0cm of is] {\texttt{retired}};
    \node (with) [rectangle, right=0cm of retired] {\texttt{with}};
    \node (the) [rectangle, right=0cm of with] {\texttt{the}};
    \node (United) [rectangle, right=0cm of the] {\texttt{United}};
    \node (States) [rectangle, right=0cm of United] {\texttt{States}};
    \node (army) [rectangle, right=0cm of States] {\texttt{army.}};
    
    \draw(he.south west)--node[yshift=-0.25cm]{ \textsc{Per}} (he.south east);
    \draw(United.south west)--node[yshift=-0.25cm]{ \textsc{Gpe}} (States.south east);
    \draw ([yshift=-0.6cm]the.south west)--node[yshift=-0.25cm]{ \textsc{Org}} ([yshift=-0.6cm]army.south east);
    \draw[white] ([yshift=-1.2cm]the.south west)--node[yshift=-0.25cm]{ \phantom{\textsc{Org}}} ([yshift=-1.2cm]army.south east);
\end{tikzpicture}
    \end{minipage}%
    \begin{minipage}[t]{0.43\textwidth}
        \vspace{-0.5cm}    \begin{tikzpicture}[
    every node/.style={
        rectangle,
        text height=1.5ex,
        text depth=.25ex
    }
]
        \node (I) [rectangle] { I};
        \node [circle, above=0.1cm of I, draw=black, inner sep=2pt] {4};
        \node (am) [rectangle, right=0cm of I] {\texttt{am}};
        \node (Fabian) [rectangle, right=0cm of am] {\texttt{Fabian}};
        \node (from) [rectangle, right=0cm of Fabian] {\texttt{from}};
        \node (Bonn) [rectangle, right=0cm of from] {\texttt{Bonn,}};
        \node (Germany) [rectangle, right=0cm of Bonn] {\texttt{Germany.}};

        \draw(I.south west)--node[yshift=-0.25cm]{ \textsc{Per}} (I.south east);
        \draw([yshift=-1.2cm]Fabian.south west)--node[yshift=-0.25cm]{ \textsc{Per}} ([yshift=-1.2cm]Germany.south east);
        \draw([yshift=-0.6cm]Bonn.south west)--node[yshift=-0.25cm]{ \textsc{Gpe}} ([yshift=-0.6cm]Germany.south east);
        \draw(Germany.south west)--node[yshift=-0.25cm]{ \textsc{Gpe}} (Germany.south east);
    \end{tikzpicture}
    \end{minipage}
    \caption{Sentence examples and their associated analyses from the \textsc{Ace-2005} dataset.}
    \label{fig:examples}
\end{figure*}
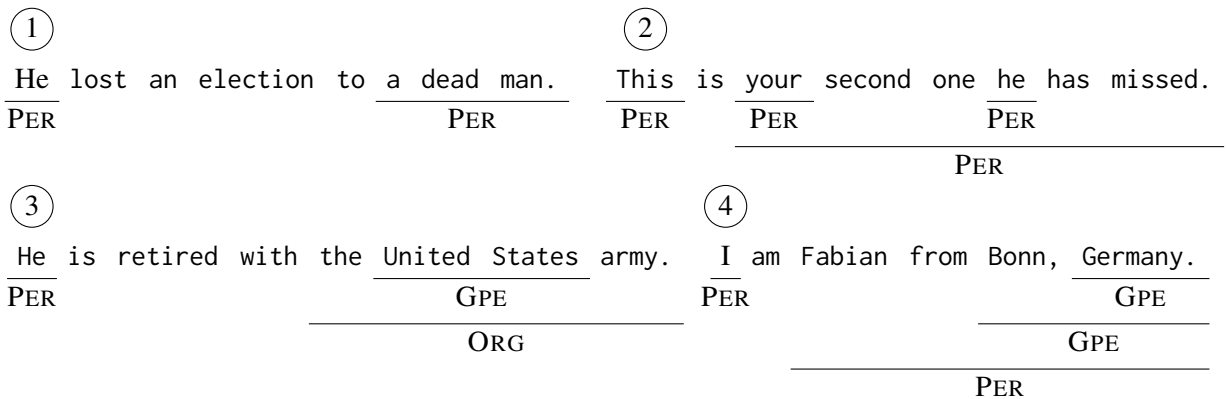

Named entity recognition (NER) is a fundamental problem in information retrieval that aims to identify mentions of entities and their associated types in natural language documents.
As such, the problem can be reduced to the identification and classification of segments of texts.
In particular, we focus on mentions that have the following properties:
\begin{enumerate}
    \item continuous, i.e.\ a mention corresponds to a contiguous sequence of words;
    \item potentially nested, i.e.\ one mention can be inside another, but they can never partially overlap.
\end{enumerate}
Four examples are shown in Figure~\ref{fig:examples}.

In a span-based setting, recognition for nested NER has a cubic-time complexity \cite{finkel2009nested,fu2021latentcyk} using variants of the Cocke-Younger-Kasami (CYK) algorithm \cite{kasami1965efficient,YOUNGER1967189,cocke1969programming}.
If we restrict the search space to non-nested mentions, then recognition can be realized in quadratic time using a semi-Markov model \cite{sarawagi2004semimarkov}.
An open question is whether it is possible to design algorithms with better time-complexity/search space trade-offs.

In this paper, we propose a novel span-based nested NER algorithm with a quadratic-time complexity, that is with the same time complexity as the semi-Markov algorithm for the non-nested case.
Our approach is based on the observation that many mentions only contain at most one nested mention of length strictly greater than one.
As such, we follow a trend in the syntactic parsing literature that studies search-spaces that allow the development of more efficient parsing algorithms, both for dependency and constituency structures \cite{pitler2012gapminding,pitler2013endpoint,satta2013headsplit,gomez2010wn,corro2020span}, \emph{inter alia}.

Our main contributions can be summarized as follows:
\begin{itemize}
    \item We present the semi-Markov and CYK-like models for non-nested and nested NER, respectively --- although we do not claim that these approaches for NER are new, our presentation of the CYK-like algorithm differs from previous work as it is tailored to the NER problem and guarantees uniqueness of derivations;
    \item We introduce a novel search space for nested NER that has no significant loss in coverage compared to the standard one, see Table~\ref{table:cover};
    \item We propose a novel quadratic-time recognition algorithm for the aforementioned search space;
    \item We experiment our quadratic-time algorithm on three English datasets (\textsc{Ace-2004}, \textsc{Ace-2005} and \textsc{Genia}) and show that it obtains comparable results to the cubic-time algorithm.
\end{itemize}

\section{Related work}

\textbf{Span-based methods:}
Semi-Markov models have been first proposed in the generative modeling framework for time-serie analysis and word segmentation \cite{janssen2013semimarkov,ge2002semimarkov}.
\citet{sarawagi2004semimarkov} introduced a discriminative variant for NER.
\citet{arora2019semi} extended this approach with a task-tailored structured SVM loss \cite{tsochantaridis2004ssvm}.
Inference algorithms for semi-Markov models have a $\mathcal{O}(n^2)$ time complexity, where $n$ is the length of the input sentence.
Unfortunately, semi-Markov models can only recognize non-nested mentions.
\citet{finkel2009nested} proposed a representation of nested mentions (together with part-of-speech tags) as a phrase structure, enabling the use of the CYK algorithm for MAP inference.
Influenced by recent work in the syntactic parsing literature on span-based model, i.e.\ models without an explicit grammar \cite{hall2014span,stern2017span}, \citet{fu2021latentcyk} proposed to rely on these span-based phrase structure parsers for nested NER.
As structures considered in NER are not \emph{stricto sensu} complete phrase-structures, they use a latent span model.
Inference in this model has a $\mathcal{O}(n^3)$ time complexity.
\citet{lou2022latentlex} extended this approach to lexicalized structures (i.e.\ where each mention has an explicitly identified head), leading to a $\mathcal{O}(n^4)$ time complexity for inference due to the richer structure.

\textbf{Tagging-based methods:}
NER can be reduced to a sentence tagging problem using BIO and BILOU schemes \cite{ratinov2009bilou} to bypass the quadratic-time complexity of semi-Markov models.
MAP Inference (resp.\ marginal inference) is then a linear-time problem using the Viterbi algorithm (resp.\ forward-backward algorithm).\footnote{It is quadratic in the number of tags, but we assume the input of the algorithm is the sentence only.}
However, this approach cannot incorporate span features neither be used for nested entities.
\citet{alex2007biomedical} and \citet{ju2018layerseq} proposed to rely on several tagging layers to predict nested entities.
\citet{shibuya2020secondbest} introduced an extension of the Viterbi algorithm that allows to rely on BIO tagging for nested NER by considering second-best paths.
To leverage the influence of outer entities, \citet{wang2021excludesecondbest} rely on different potential functions for inner entities.
Note that algorithms for the second-best paths method have a $\mathcal O(n^2)$ time complexity, that is similar to the span-based alogithm we propose.

\textbf{Hypergraph-based methods:}
\citet{lu2015hypergraph} proposed an hypergraph-based method for nested NER.
Although this approach is appealing for its $\mathcal{O}(n)$ (approximate) inference algorithms, it suffers from two major issues: (1) the marginal inference algorithm overestimate the partition function; (2) the representation is ambiguous, that is a single path in the hypergraph may represent different analysis of the same sentence.
\citet{muis2017hypergraph} proposed a different hypergraph with $\mathcal O(n^2)$ inference algorithms that solves issue (1) but still exhibits issue (2).
\citet{katiyar2018hypergraph} extended hypergraph methods to rely on neural network scoring.
\citet{wang2018hypergraph} proposed a novel hypergraph method that fixes issues (1) and (2) but their approach does not forbid partially overlapping mentions.

\textbf{Unstructured methods:}
Several authors proposed to predict the presence of a mention on each span independently, sometimes with specialized neural architectures \cite{xu2017local,sohrab2018exhaustive,zheng2019boundary,xia2019multi,wang2020pyramid,tan2020boundary,zaratiana2022gnner}, \emph{inter alia}.
Note that these approaches classify $\mathcal O(n^2)$ spans of text independently, hence  the time-complexity is similar to the approach proposed in this paper but they cannot guarantee well-formedness of the prediction.
\section{Nested named-entity recognition}

In this section, we introduce the nested NER problem and the vocabulary we use through the paper.

\subsection{Notations and vocabulary}

Let $\vs = \evs_1 ... \evs_n$ be a sentence of $n$ words.
Without loss of generality, we assume that all sentences are of the same size.
We use interstice (or fencepost) notation to refer to spans of $\vs$, i.e.\ $\vs_{i:j} = \evs_{i+1} ... \evs_{j}$ if $0 \leq i < j \leq n$, the empty sequence if $0 \leq i = j \leq n$ and undefined otherwise.
We denote $M$ the set of possible mentions in a sentence and $T$ the set of mention types, e.g.\ $T = \{ \textsc{Per}, \textsc{Org}, \textsc{Gpe}, ...\}$.
Without loss of generality, we assume that $T \cap \{\rightarrow, \leftrightarrow, \mapsto, \mapsfrom \} = \emptyset$.
A mention is denoted $\langle t, i, j \rangle \in M$ s.t.\ $t \in T, 0 \leq i < j \leq n$, where $i$ (resp.\ $j$) is called the \textbf{left border} (resp.\ \textbf{right border}).
An analysis of sentence $\vs$ is denoted $\vy \in \{0, 1\}^M$ where $\evy_m = 1$ (resp.\ $\evy_m = 0$) indicates that mention $m \in M$ is included in the analysis (resp.\ is not included).
For example, the analysis of sentence 1 in Figure~\ref{fig:examples} is represented by a vector $\vy$ where $\evy_{\langle \textsc{Per}, 0, 1\rangle} = 1$, $\evy_{\langle \textsc{Per}, 5, 8\rangle} = 1$ and all other elements are equal to zero.
A mention $\langle t, i, j \rangle$ is said to be inside another mention $\langle t', i', j'\rangle$ iff $i' < i < j \leq j'$ or $i' \leq i < j < j'$.

Let $\vy$ be the analysis of a sentence.
We call \textbf{first level mentions} all mentions in $\vy$ that are not inside another mention of the analysis.
We call \textbf{nested mentions} all mentions that are not first level mentions.
For example, the first level mentions of the analysis of sentence 2 in Figure~\ref{fig:examples} are $\langle \textsc{Per}, 0, 1 \rangle$ ``\texttt{this}'' and $\langle \textsc{Per}, 2, 8 \rangle$ ``\texttt{your second one he has missed}''.
We call \textbf{children} of mention $m \in M$ the set $C \subseteq M$ of mentions that are inside $m$ but not inside another mention that is inside $m$.
Conversely, $m$ is said to be the parent of each mention in $C$.
For example, in sentence 2 in Figure~\ref{fig:examples}, the mention $\langle \textsc{Per}, 2, 8 \rangle$ ``\texttt{your second one he has missed}'' has two children, $\langle \textsc{Per}, 2, 3 \rangle$ ``\texttt{your}'' and $\langle \textsc{Per}, 5, 6 \rangle$ ``\texttt{he}''.
In sentence 4 in Figure~\ref{fig:examples}, $\langle \textsc{Gep}, 5, 6 \rangle$ ``\texttt{Germany}'' is a child of $\langle \textsc{Gep}, 4, 6 \rangle$ ``\texttt{Bonn, Germany}'' but it is not a child of $\langle \textsc{Per}, 2, 6 \rangle$``\texttt{Fabian from Bonn, Germany}''.
The \textbf{left neighborhood} (resp.\ \textbf{right neighborhood}) of a nested mention is the span between the left  border of its parent and its left border (resp.\ between its right border and the right border of its parent).
For example, in sentence 2 in Figure~\ref{fig:examples}, mention $\langle \textsc{Per}, 5, 6 \rangle$ ``\texttt{he}'' has left neighborhood $\vs_{2:5}$ ``\texttt{your second one}'' and  right neighborhood $\vs_{6,8}$ ``\texttt{has missed}''.

The set of possible analyses is denoted $Y$.
We will consider three different definitions of $Y$:
\begin{enumerate}
    \item the set of analyses where no disjoint mention spans overlap, corresponding to non-nested NER;
    \item the set of analyses where one mention span can be inside another one but cannot partially overlap, corresponding to nested NER;
    \item the set 2 with additional constraint that a mention must contain at most one child with a span length strictly greater to one.
\end{enumerate}

\subsection{Inference problems}

The weight of an analysis $\vy \in Y$ is defined as the sum of included mention weights.
Let $\vw \in \R^{M}$ be a vector of mention weights.
The probability of an analysis is defined via the Boltzmann or ``softmax'' distributions:
\begin{align*}
p(\vy | \vw) &= \frac{\exp(\vw^\top \vy)}{Z(\vw)},
\end{align*}
where $Z(\vw) = \sum_{\vy' \in Y} \exp(\vw^\top \vy')$ is the partition function.
Note that, in general, the set $Y$ is of exponential size but $Z(\vw)$ can nonetheless be efficiently computed via dynamic programming.

The training problem aims to minimize a loss function over the training data.
We focus on the negative log-likelihood loss function defined as:
$$
\ell(\vw, \vy) = - \vw^\top \vy + \log Z(\vw).
$$
Note that this loss function is convex in $\vw$.
This differentiates us from previous work that had to rely on non-convex losses \cite{fu2021latentcyk,lou2022latentlex}.
Moreover, note that the loss function used by \citet{fu2021latentcyk} and \citet{lou2022latentlex} requires to compute the log-partition twice, one time with ``normal'' weights and one time with masked weights.
The difference lays in the fact that we will use algorithms that are tailored for the considered search space $Y$ whereas \citet{fu2021latentcyk} and \citet{lou2022latentlex} introduced latent variables in order to be able to rely on algorithms designed for a different problem, namely syntactic constituency parsing.
Note that the partial derivatives of $\log Z(\vw)$ are the marginal distributions of mentions \cite{wainwright2008exp}.
Hence, we will refer to computing $\log Z(\vw)$ and its derivatives as marginal inference, a required step for gradient based optimization at training time.

At test time, we aim to compute the highest scoring structure given weights $\vw$:
\begin{align*}
    \vy^* = \argmax_{\vy \in Y} p(\vy | \vw)
    = \argmax_{\vy \in Y} \vw^\top\vy
\end{align*}
We call this problem MAP inference.

For many problems in natural language processing, marginal inference and MAP inference can be computed via dynamic programming over different semirings \cite{goodman1999semiring} or dynamic programming with smoothed max operators \cite{mensch18ddp}.
However, we need to ensure the uniqueness of derivations property so that a single analysis $\vy \in Y$ has exactly one possible derivation under the algorithm.
Otherwise, the same analysis would be counted several times when computing the partition function, leading to an overestimation of its value.

\begin{table*}[p]
    \centering
    \small

\begin{minipage}[t]{0.47\textwidth}
    \centering
    \begin{tabular}{@{}r@{\hskip 0.1in}l@{\hskip 0.15in}l@{\hskip 0.15in}l@{}}
     \toprule
     &  Items & Rules & Comments\\
     \midrule
     1. & $[\rightarrow, 0]$ & Axiom & Recognize mention \\
     2. & $[\textsc{Per}, 0, 1]$ & Axiom & ``He''  \\
     3. & $[\rightarrow, 1]$ & \textbf{(a)} with 1 and 2 &  \\
     \midrule
     4. & $[\rightarrow, 2]$ & \textbf{(b)} with 3 & Recognize empty  \\
     5. & $[\rightarrow, 3]$ & \textbf{(b)} with 4 & space between  \\
     6. & $[\rightarrow, 4]$ & \textbf{(b)} with 5 & the two mentions\\
     7. & $[\rightarrow, 5]$ & \textbf{(b)} with 6 \\
     \midrule
     8. & $[\textsc{Per}, 5, 8]$ & Axiom & Recognize mention \\
     9. & $[\rightarrow, 8]$ & \textbf{(a)} with 7 and 8 & ``a dead man'' \\
     \bottomrule
\end{tabular}

    \captionof{table}{Example of recognition trace with the semi-Markov algorithm on sentence 1.}
    \label{tab:alg:semimarkov}
    \vspace{0.5cm}
    \begin{tabular}{@{}r@{\hskip 0.1in}l@{\hskip 0.15in}l@{\hskip 0.15in}l@{\hskip 0.05in}l@{}}
     \toprule
     &  Items & Rules $\mathcal O(n^3)$& $\mathcal O(n^2)$ & Comments\\
     \midrule
     1. & $[\mapsto, 0, 0]$ & Axiom & // & Recognize  \\
     2. & $[\mapsto, 0, 1]$ & \textbf{(d)} with 1 & // & mention\\
     3. & $[\textsc{Per}, 0, 1]$ & Axiom & // & ``This''\\
     4. & $[\leftrightarrow, 0, 1]$ & \textbf{(g)} with 2 \& 3 & // \\
     \midrule
     5. & $[\mapsto, 2, 2]$ & Axiom & //& Recognize  \\
     6. & $[\mapsto, 2, 3]$ & \textbf{(d)} with 5 & // & mention\\
     7. & $[\textsc{Per}, 2, 3]$ & Axiom & // & ``your''\\
     8. & $[\leftrightarrow, 2, 3]$ & \textbf{(g)} with 7& // \\
     \midrule
     9. & $[\mapsto, 0, 5]$ & Axiom & //& Recognize  \\
     10. & $[\mapsto, 0, 6]$ & \textbf{(d)} with 9 & // & mention \\
     11. & $[\textsc{Per}, 5, 6]$ & Axiom & // & ``he''\\
     12. & $[\leftrightarrow, 5, 6]$ & \textbf{(g)} with 11 & //\\
     \midrule
     13. & $[\mapsto, 2, 4]$ & \textbf{(f)} with 8 & //& Recognize \\
     14. &$[\mapsto, 2, 5]$ & \textbf{(d)} with 13 & //&mention  \\
     15. &$[\mapsto, 2, 6]$ & \textbf{(c)} with 14 \& 12& \textbf{(j)} & ``your second  \\
     16. & $[\mapsto, 2, 7]$ & \textbf{(d)} with 15 & //& one he has \\
     17. & $[\mapsto, 2, 8]$ & \textbf{(d)} with 16 & //& missed''\\
     18. & $[\textsc{Per}, 2, 8]$ & Axiom & //\\
     19. & $[\leftrightarrow, 2, 8]$ & \textbf{(g)} with 17 \& 18& // \\
     \midrule
     20. & $[\rightarrow, 0]$ & Axiom & //& Combine all   \\
     21. & $[\rightarrow, 1]$ &\textbf{(h)} with 20 \& 4& // & first-level   \\
     22. & $[\rightarrow, 2]$ &\textbf{(i)} with 21& // & mentions \\
     23. & $[\rightarrow, 8]$ &\textbf{(h)} with 22 \& 19& // & \\
     \bottomrule
\end{tabular}
\captionof{table}{Example of recognition trace with the CYK-like and the proposed $\mathcal{O}(n^2)$ algorithm on sentence 2. There is only one rule that differs, but they both share the same antecedents.}
    \label{tab:alg:cyk}
\end{minipage}\hfill%
\begin{minipage}[t]{0.47\textwidth}
    \centering
    \begin{tabular}{@{}r@{\hskip 0.1in}l@{\hskip 0.15in}l@{\hskip 0.15in}l@{}}
\toprule
&  Items & Rules & Comments\\
\midrule
1. & $[\mapsto, 0, 0]$ & Axiom & Recognize  \\
2. & $[\mapsto, 0, 1]$ & \textbf{(d)} with 1 & mention\\
3. & $[\textsc{Per}, 0, 1]$ & Axiom & ``He''\\
4. & $[\leftrightarrow, 0, 1]$ & \textbf{(g)} with 2 and 3 \\
\midrule
5. & $[\mapsto, 5, 5]$ & Axiom & Recognize  \\
6. & $[\mapsto, 5, 6]$ & \textbf{(d)} with 5 & mention\\
7. & $[\mapsto, 5, 7]$ & \textbf{(d)} with 6 & ``United states''\\
8. & $[\textsc{Gpe}, 5, 7]$ & Axiom \\
9. & $[\leftrightarrow, 5, 7]$ & \textbf{(a)} with 1 and 2 \\
\midrule
10. & $[\mapsfrom, 4, 7]$ & \textbf{(m)} with 9 & Recognize   \\
11. & $[\mapsto, 4, 7]$ & \textbf{(p)} with 10 & mention \\
11. & $[\mapsto, 4, 8]$ & \textbf{(d)} with 11 & ``the United  \\
12. & $[\textsc{Org}, 4, 8]$ & Axiom & States army''\\
13. & $[\leftrightarrow, 4, 8]$ & \textbf{(g)} with 11 and 12 \\
\midrule
14. & $[\rightarrow, 0]$ & Axiom & Combine all  \\
15. & $[\rightarrow, 1]$ & \textbf{(h)} with 14 and 4 & first-level  \\
16. & $[\rightarrow, 2]$ & \textbf{(i)} with 15 & mentions \\
17. & $[\rightarrow, 3]$ & \textbf{(i)} with 16 \\
18. & $[\rightarrow, 4]$ & \textbf{(i)} with 17 \\
19. & $[\rightarrow, 8]$ & \textbf{(h)} with 18 \& 13 \\
\bottomrule
\end{tabular}
\captionof{table}{Example of recognition trace of the proposed algorithm on sentence 3.}
    \label{tab:alg:quadratic1}
    \vspace{0.5cm}

    \begin{tabular}{@{}r@{\hskip 0.1in}l@{\hskip 0.15in}l@{\hskip 0.15in}l@{}}
\toprule
&  Items & Rules & Comments\\
\midrule
1. & $[\mapsto, 0, 0]$ & Axiom & Recognize  \\
2. & $[\mapsto, 0, 1]$ & \textbf{(d)} with 1 &mention\\
3. & $[\textsc{Per}, 0, 1]$ & Axiom & ``I''\\
4. & $[\leftrightarrow, 0, 1]$ & \textbf{(g)} with 2 \& 3 \\
\midrule
5. & $[\mapsto, 5, 5]$ & Axiom & Recognize  \\
6. & $[\mapsto, 5, 6]$ & \textbf{(d)} with 5 & mention\\
7. & $[\textsc{Gpe}, 5, 6]$ & Axiom & ``Germany'' \\
8. & $[\leftrightarrow, 5, 6]$ & \textbf{(g)} with 6 \& 7 \\
\midrule
9. & $[\mapsto, 4, 4]$ & Axiom & Recognize   \\
10. & $[\mapsto, 4, 5]$ & \textbf{(d)} with 9 & mention\\
11. & $[\mapsto, 4, 6]$ & \textbf{(j)} with 10 \& 8 & ``Bonn, Germany'' \\
12. & $[\textsc{Gpe}, 4, 6]$ & Axiom \\
13. & $[\leftrightarrow, 4, 6]$ & \textbf{(g)} with 11 \& 12 \\
\midrule
14. & $[\mapsfrom, 3, 6]$ & \textbf{(m)} with 13 & Recognize   \\
15. & $[\mapsfrom, 2, 6]$ & \textbf{(n)} with 14 & mention  \\
16. & $[\mapsto, 2, 6]$ & \textbf{(p)} with 15 & `Fabian from\\
17. & $[\textsc{Per}, 2, 6]$ & Axiom & Bonn, Germany'' \\
18. & $[\leftrightarrow, 2, 6]$ & \textbf{(g)} with 16 \& 17 \\
\midrule
19. & $[\rightarrow, 0]$ & Axiom & Combine all  \\
20. & $[\rightarrow, 1]$ & \textbf{(h)} with 19 \& 4 & first-level  \\
21. & $[\rightarrow, 2]$ & \textbf{(i)} with 10 & mentions\\
22. & $[\rightarrow, 6]$ & \textbf{(h)} with 21 \& 18 \\
\bottomrule
\end{tabular}
\captionof{table}{Example of recognition trace of the proposed algorithm on sentence 4.}
    \label{tab:alg:quadratic2}
\end{minipage}

\end{table*}
\section{Related algorithms}

In this section, we present semi-Markov and CYK-like algorithms for non-nested and nested NER, respectively.
Our presentation is based on the weighted logic programming formalism, also known as parsing-as-deduction \cite{pereira1983parsing}.
We refer the reader to \citet[][Chapter 3]{kallmeyer2010parsing} for an introduction to this formalism.
The space and time complexities can be directly inferred by counting the maximum number of free variables in items and deduction rules, respectively.
To the best of our knowledge, the presentation of the CYK-like algorithm is novel as previous work relied on the ``actual'' CYK algorithm \cite{finkel2009nested} or its variant for span-based syntactic parsing \cite{lou2022latentlex,fu2021latentcyk}.

\subsection{Non-nested named-entity recognition}
\label{sec:semimarkov}

The semi-Markov algorithm recognizes a sentence from left to right.
Items are of the following forms:
\begin{itemize}
    \item $[t, i, j]$ s.t.\ $t \in T$ and $0 \leq i < j \leq n$: represent the mention $\langle t, i, j\rangle$;
    \item $[\rightarrow, i]$ s.t.\ $0 \leq i \leq n$: represent a partial analysis of the sentence covering words $\vs_{0:i}$.
\end{itemize}
Axioms are items of the form $[\rightarrow, 0]$ and $[t, i, j]$.
The first axiom form represents an empty partial analysis and the second set of axioms represent all possible mentions in the sentence.
We assign weight $\evw_{\langle t, i, j \rangle}$ to axiom $[t, i, j]$, for all $t \in T, i, j \in \mathbb N \text{ s.t.\ } 0 \leq i < j \leq n$.
The goal of the algorithm is the item $[\rightarrow, n]$.

Deduction rules are defined as follows:\newline
\begin{minipage}{\dimexpr.25\textwidth-.5\columnsep}
    \begin{prooftree}
        \LeftLabel{ \textbf{(a)}}
    	\AxiomC{$[\rightarrow, i]$}
    	\AxiomC{$[t, i, j]$}
    	\BinaryInfC{$[\rightarrow, j]$}
    \end{prooftree}
\end{minipage}%
\begin{minipage}{\dimexpr.25\textwidth-.5\columnsep}
    \begin{prooftree}
        \LeftLabel{ \textbf{(b)}}
    	\AxiomC{$[\rightarrow, i-1]$}
    	\UnaryInfC{$[\rightarrow, i]$}
    \end{prooftree}
\end{minipage}\newline
Rule \textbf{(a)} appends a mention spanning words $\vs_{i:j}$ to a partial analysis, whereas rule \textbf{(b)} advances one position by assuming word $\vs_{i:i+1}$ is not covered by a mention.

A trace example of the algorithm is given in Table~\ref{tab:alg:semimarkov}.
Soundness, completeness and uniqueness of derivations can be directly induced from the deduction system.
The time and space complexities are both $\mathcal O(n^2|T|)$.

\subsection{Nested named-entity recognition}
\label{sec:cyk}

We present a CYK-like algorithm for nested named entity recognition.
Contrary to algorithms proposed by \citet{finkel2009nested} and \citet{fu2021latentcyk}, \emph{inter alia}, our algorithm directly recognizes the nested mentions and does not require any ``trick'' to take into account non-binary structures, words that are not covered by any mention or the fact that a word in a mention may not be covered by any of its children.
As such, we present an algorithm that is tailored for NER instead of the usual ``hijacking'' of constituency parsing algorithms.
This particular presentation of the algorithm will allow us to simplify the presentation of our novel contribution in Section~\ref{sec:algorithm}.

Items are of the following forms:
\begin{itemize}
    \item $[t, i, j]$ as defined previously;
    \item $[\rightarrow, i]$ as defined previously;
    \item $[\mapsto, i, j]$ with $0 \leq i < j \leq n$: represent the partial analysis of a mention and its nested structure starting at position $i$.
    \item $[\leftrightarrow, i, j]$ with $0 \leq i < j \leq n$: represent the full analysis of a mention spanning $\vs_{i: j}$, including its internal structure (i.e.\ full analysis of its children).
\end{itemize}
Axioms and goals are the same as the ones of the semi-Markov algorithm presented in Section~\ref{sec:semimarkov}, with supplementary set of items of form $[\mapsto, i, i]$ that are used to start recognizing the internal structure of a mention starting at position $i$.

The algorithm consists of two steps.
First, the internal structure of mentions are constructed in a bottom-up fashion.
Second, first level mentions (and their internal structures) are recognized in a similar fashion to the semi-Markov model.
The deduction rules for bottom-up construction are defined as follows:\newline
\begin{minipage}{\dimexpr.30\textwidth-.5\columnsep}
\begin{prooftree}
    \LeftLabel{\textbf{(c)}}
    \RightLabel{\small i < k}
    \AxiomC{$[\mapsto, i, k]$}
    \AxiomC{$[\leftrightarrow, k, j]$}
    \BinaryInfC{$[\mapsto, i, j]$}
\end{prooftree}
\end{minipage}%
\begin{minipage}{\dimexpr.20\textwidth-.5\columnsep}
    \begin{prooftree}
        \LeftLabel{~~~~\textbf{(d)}}
        \AxiomC{$[\mapsto, i, j-1]$}
        \UnaryInfC{$[\mapsto, i, j]$}
    \end{prooftree}
\end{minipage}
\begin{minipage}{\dimexpr.26\textwidth-.5\columnsep}
\begin{prooftree}
    \LeftLabel{\textbf{(e)}}
    \AxiomC{$[\leftrightarrow, i, k]$}
    \AxiomC{$[\leftrightarrow, k, j]$}
    \BinaryInfC{$[\mapsto, i, j]$}
\end{prooftree}
\end{minipage}%
\begin{minipage}{\dimexpr.24\textwidth-.5\columnsep}
\begin{prooftree}
    \LeftLabel{\textbf{(f)}}
    \AxiomC{$[\leftrightarrow, i, j-1]$}
    \UnaryInfC{$[\mapsto, i, j]$}
\end{prooftree}
\end{minipage}
\begin{prooftree}
    \LeftLabel{\textbf{(g)}}
    \RightLabel{\small i < j}
    \AxiomC{$[\mapsto, i, j]$}
    \AxiomC{$[t, i, j]$}
    \BinaryInfC{$[\leftrightarrow, i, j]$}
\end{prooftree}
Rule~\textbf{(c)} concatenates an analyzed mention to a partial analysis of another mention --- note that the constraint forbids that right antecedent shares its left border with its parent.
Rule~\textbf{(d)} advances of one position in the partial structure, assuming the analyzed mention starting at $i$ does not have a child mention covering $\vs_{j-1:j}$.
Rules~\textbf{(e)} and~\textbf{(f)} are used to recognize the internal structure of a mention that has a child sharing the same left border.
Although the latter two deduction rules may seem far-fetched, they cannot be simplified without breaking the uniqueness of derivations property or breaking the prohibition of self loop construction of $\leftrightarrow$ items.
Finally, rule~\textbf{(g)} finishes the analysis of a mention and its internal structure.

Note that this construction is highly similar to the dotted rule construction in the Earley algorithm \cite{earley1970efficient}.
Moreover, contrary to \citet{stern2017span}, we do not introduce null labels for implicit binarization.
The benefit of our approach is that there is no spurious ambiguity in the algorithm,
i.e.\ we guaranty uniqueness of derivations.
Therefore, we can use the same deduction rules to compute the log-partition function of the negative log-likelihood loss.
This is not the case of the approach of \citet{stern2017span}, which forces them to rely on a structured hinge loss.

Deduction rules for the second step are defined as follows:
\begin{minipage}{\dimexpr.3\textwidth-.5\columnsep}
    \begin{prooftree}
        \LeftLabel{\textbf{(h)}}
    	\AxiomC{$[\rightarrow, i]$}
    	\AxiomC{$[\leftrightarrow, i, j]$}
    	\BinaryInfC{$[\rightarrow, j]$}
    \end{prooftree}
\end{minipage}%
\begin{minipage}{\dimexpr.2\textwidth-.5\columnsep}
    \begin{prooftree}
        \LeftLabel{\textbf{(i)}}
    	\AxiomC{$[\rightarrow, i-1]$}
    	\UnaryInfC{$[\rightarrow, i]$}
    \end{prooftree}
\end{minipage}\newline
They have similar interpretation to the rules of the semi-Markov model where we replaced mentions by possibly nested structures.

A trace example of the algorithm is given in Table~\ref{tab:alg:cyk}.
Although the algorithm is more involved than usual presentations, our approach directly maps a derivation to nested mentions and guarantee uniqueness of derivations.
The space and time complexities are $\mathcal{O}(n^2|T|)$ and $\mathcal{O}(n^3|T|)$, respectively.
\section{$\mathcal{O}(n^2)$ nested named-entity recognition}
\label{sec:algorithm}

In this section, we describe our novel algorithm for quadratic-time nested named entity recognition.
Our algorithm limits its search space to mentions that contain at most one child of length strictly greater to one.

Items are of the following forms:
\begin{itemize}
    \item $[t, i, j]$ as defined previously;
    \item $[\rightarrow, i]$ as defined previously;
    \item $[\mapsto, i, j]$ as defined previously;
    \item $[\leftrightarrow, i, j]$ as defined previously;
    \item $[\mapsfrom, i, j]$ with $0 \leq i < j \leq 0$: represents a partial analysis of a mention and its internal structure, where its content will be recognized by appending content on the left instead of the right.
\end{itemize}
Axioms and goals are the same than the one of the CYK-like algorithm presented in Section~\ref{sec:cyk} --- importantly, there is no extra axiom for items of the form $[\mapsfrom, i, j]$.

For the moment, assume we restrict nested mentions that have a length strictly greater to the ones that share their left boundaries with their parent.
We can re-use rules~\textbf{(d)}, \textbf{(f)}, \textbf{(g)}, \textbf{(h)} and~\textbf{(i)} together with the following two deduction rules:
\begin{prooftree}
    \LeftLabel{\textbf{(j)}}
    \RightLabel{\small $i < j - 1$}
    \AxiomC{$[\mapsto, i, j-1]$}
    \AxiomC{$[\leftrightarrow, j-1, j]$}
    \BinaryInfC{$[\mapsto, i, j]$}
\end{prooftree}
\begin{prooftree}
    \LeftLabel{\textbf{(k)}}
    \AxiomC{$[\leftrightarrow, i, j-1]$}
    \AxiomC{$[\leftrightarrow, j-1, j]$}
    \BinaryInfC{$[\mapsto, i, j]$}
\end{prooftree}
More precisely, we removed the two rules inducing a cubic-time complexity in the CYK-like algorithm and replaced them with quadratic-time rules.
This transformation is possible because our search space forces the rightmost antecedents of these two rules to cover a single word, hence we do not need to introduce an extra free variable.
However, in this form, the algorithm only allows the child mention of length strictly greater to one to share its left boundary with its parent.

We now extend the algorithm to the full targeted search space.
The intuition is as follows: for a given mention, if it has a child mention of length strictly greater than one that does not share its left border with its parent, we first start recognizing this child mention and its left neighborhood and then move to right neighborhood using previously defined rules.
We start the recognition of the left neighborhood using the two following rules:
\begin{prooftree}
    \LeftLabel{\textbf{(l)}}
    \RightLabel{$i + 2 < j$}
    \AxiomC{$[\leftrightarrow, i, i+1]$}
    \AxiomC{$[\leftrightarrow, i+1, j]$}
    \BinaryInfC{$[\mapsfrom, i, j]$}
\end{prooftree}
\begin{prooftree}
    \LeftLabel{\textbf{(m)}}
    \RightLabel{$i + 2 < j$}
    \AxiomC{$[\leftrightarrow, i+1, j]$}
    \UnaryInfC{$[\mapsfrom, i, j]$}
\end{prooftree}
where the constraints ensure antecedents $[\leftrightarrow, i+1, j]$ are non-unary (otherwise we will break the uniqueness of derivations constraint).
Rule \textbf{(l)} (resp.\ \textbf{(m)}) recognizes the case where span $\vs_{i:i+1}$ contains (resp.\ does not contain) a mention.
The following rules are analogous to rules $\textbf{(d)}$ and $\textbf{(j)}$ but for visiting the left neighborhood instead of the right one:
\begin{minipage}{0.17\textwidth}
\begin{prooftree}
    \LeftLabel{\textbf{(n)}}
    \AxiomC{$[\mapsfrom, i+1, j]$}
    \UnaryInfC{$[\mapsfrom, i, j]$}
\end{prooftree}
\end{minipage}%
\begin{minipage}{0.33\textwidth}
    \begin{prooftree}
        \LeftLabel{\textbf{(o)}}
        \AxiomC{$[\leftrightarrow, i, i+1]$}
        \AxiomC{$[\mapsfrom, i+1, j]$}
        \BinaryInfC{$[\mapsfrom, i, j]$}
    \end{prooftree}
\end{minipage}\newline
Finally, once the left neighborhood has been recognized, we move to the right one using the following rule:
\vspace{-1em}\begin{prooftree}
    \LeftLabel{\textbf{(p)}}
    \AxiomC{$[\mapsfrom, i, j]$}
    \UnaryInfC{$[\mapsto, i, j]$}
\end{prooftree}

Using the aforementioned rules, our algorithm has time and space complexities of $\mathcal O(n^2 |T|)$.
We illustrate the difference with the CYK-like algorithm with a trace example in Table~\ref{tab:alg:cyk}: in this specific example, the two analyses differ only by the application of a single rule.
Table~\ref{tab:alg:quadratic1} contains a trace example where all nested mentions have a size one, so the parent mention is visited from left to right.
Table~\ref{tab:alg:quadratic2} contains a trace example where we need to construct one internal structure by visiting the left neighborhood of the non-unary child mention from right to left.

Soundness and completeness can be proved by observing that, for a given mention, any children composition can be parsed with deduction rules as long as there is at most one child with a span strictly greater to one.
Moreover, these are the only children composition that can be recognized.
Finally, uniqueness of derivations can be proved as there is a single construction order of the internal structure of a mention.

\textbf{Infinite recursion.}
An important property of our algorithm is that it does not bound the number of allowed recursively nested mentions.
For example, consider the phrase ``\emph{[Chair of [the Committee of [Ministers of [the Council of [Europe]]]]]}''.
Not only can this nested mention structure be recognized by our algorithm, but any supplementary ``of'' precision would also be recognized.

\textbf{Possible extension.}
Note that we could extend the algorithm so that we allow each mention to have at most one child of length strictly greater to a predefined constant $c$, and other children should have a length less or equal to $c$.
However, as fixing $c = 1$ results in a good cover of datasets we use, we do not consider this extension in this work.

\begin{table}[t]
    \centering
    \small
    \begin{tabular}{@{}lccc@{}}
        \toprule
         &  \textsc{Ace-2004} & \textsc{Ace-2005} &\textsc{Genia}\\
         \midrule
        Non-nested $\mathcal{O}(n^2)$ & 78.19  & 80.89 & 91.21  \\
        Nested $\mathcal{O}(n^3)$ & 99.97  & 99.96  & 99.95  \\
        Nested $\mathcal{O}(n^2)$ & 98.92  & 99.31  & 99.83 \\
        \bottomrule
    \end{tabular}
    \caption{Maximum recall that can be achieved on the full datasets (train, dev and test) for the three algorithms.}
    \label{table:cover}
\end{table}
\begin{figure}[t]
    \centering
		\begin{tikzpicture}
		\begin{axis}[
		width=6cm,
		height=5cm,
		xlabel={\small Sentence length},
		ylabel={\small Seconds per sentence},
		xmin=0, xmax=300,
		ymin=0, ymax=0.005,
		tick style={color=white},
		xtick={50,150,250},
		ytick={0, 0.0025, 0.005},
		legend pos=north west,
		axis lines=left,
		yticklabel style={
                /pgf/number format/fixed,
                /pgf/number format/precision=5
        },
        scaled y ticks=false,
        tick label style={font=\tiny}
		]
		
		\addplot[black, solid, thick] table[col sep=comma, x=x, y=y] {timing/timing_cfg};
		\addplot[black, dashed, thick] table[col sep=comma, x=x, y=y] {timing/timing_semimarkov};
		\addplot[black, dotted, thick] table[col sep=comma, x=x, y=y] {timing/timing_single};
		
		\end{axis}
		\end{tikzpicture}
		\caption{%
		    MAP decoding time (dynamic programming algorithm only) in seconds on an Intel Core i5 (2.4~GHz) processor for sentences of lengths 1 to 300.
		    \textbf{(dashed)}~quadratic-time semi-markov algorithm.
		    \textbf{(solid)}~CYK-like cubic-time algorithm.
		    \textbf{(dotted)}~quadratic-time algorithm proposed in this paper.
		}
    \label{fig:time}
\end{figure}
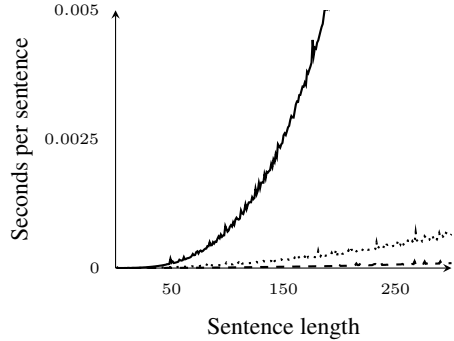
\begin{table*}[t]
    \centering
    \small
    \begin{tabular}{@{}l@{\hskip 0.3in}ccc@{\hskip 0.3in}ccc@{\hskip 0.3in}ccc@{}}
        \toprule
         & \multicolumn{9}{c}{Dataset} \\
         \cmidrule{2-10}
         & \multicolumn{3}{c}{\textsc{Ace-2004}} & \multicolumn{3}{c}{\textsc{Ace-2005}} & \multicolumn{3}{c}{\textsc{Genia}} \\
         & Prec. & Rec. & F1 & Prec. & Rec. & F1 & Prec. & Rec. & F1 \\
         \midrule
         \multicolumn{10}{@{}l@{}}{\textbf{Comparable models based on \textsc{BERT}}} \\
         \midrule
         \citet{shibuya2020secondbest} & 85.23 & 84.72 & 84.97 & 83.30 & 84.69 & 83.99 & 77.46 & 76.65 & 77.05\\
         %
         \citet{wang2020pyramid} & 86.08 & 86.48 & 86.28 & 83.95 & 85.39 & 84.66 & 79.45 & 78.94 & 79.19 \\
         %
         \citet[][max]{wang2021excludesecondbest} & 86.27 & 85.09 & 85.68 & 85.28 & 84.15 & 84.71 &  79.20 & 78.16 & 78.67\\
         \citet{fu2021latentcyk}\textsuperscript{\textdagger} & 87.62 & 87.57 & 87.60 & 83.34 & 85.67 & 84.49 & 79.10 & 76.53 & 77.80 \\
         \citet{tan2021seqtoset}\textsuperscript{\textdagger} & 87.05 & 86.26 & 86.65 & 83.92 & 84.75 & 84.33 & 78.33 & 76.66 & 77.48 \\
         \citet{shen2021locate}\textsuperscript{\textdagger} & 87.27 & 86.61 & 86.94 & 86.02 & 85.62 & 85.82 & 76.80 & 79.02 & 77.89 \\
         \citet{yan2021unifiedgen} & 87.27 & 86.41 & 86.84 & 83.16 & 86.38 & 84.74 & 78.57 & 79.30 & 78.93 \\
         \midrule
         \multicolumn{10}{@{}l@{}}{\textbf{Model based on \textsc{Bert} with lexicalization}} \\
         \midrule
         \citet{lou2022latentlex} & 87.39 & 88.40 & 87.90 & 85.97 & 87.87 & 86.91 & 78.39 & 78.50 & 78.44 \\
         \midrule
         \multicolumn{10}{@{}l@{}}{\textbf{This work}} \\
         \midrule
         Semi-Markov algorithm, $\mathcal O(n^2)$
         & 89.06 & 68.63 & 77.52
         & 84.39 & 68.51 & 75.63
         & 80.87 & 71.37 & 75.82
         \\
         CYK-like algorithm, $\mathcal O(n^3)$ 
          & 87.18 & 86.06 & 86.62 
         & 84.20 & 85.84 & 85.01 
         & 79.20 & 77.31 & 78.24 
         \\

         Proposed algorithm, $\mathcal O(n^2)$ 
                   & 87.37 & 85.04 & 86.19  
         & 84.42 & 85.28 & 84.85  
         & 79.28 & 77.25 & 78.25 
         \\
         \bottomrule
    \end{tabular}
    \caption{Precision, recall and F1-measure results. We compare ourselves to other BERT-based models --- some of the cited papers includes richer models that we omit for brievity as our goal is only to asses the performance of our algorithm compared to the CYK-like one. Results marked with \textdagger{} are the reproduction of \citet{lou2022latentlex} as the original papers experimented on different data splits.}
    \label{tab:results}
\end{table*}

\section{Experimental results}

\textbf{Data.}
We evaluate our algorithms on the \textsc{Ace-2004} \cite{ace2004}, \textsc{Ace-2005} \cite{ace2005} and \textsc{Genia} \cite{genia} datasets.
We split and pre-process the data using the tools distributed by \citet{shibuya2020secondbest}.

\textbf{Data coverage.}
As our parsing algorithm considers a restricted search space, an important question is whether it has a good coverage of NER datasets.
Table~\ref{table:cover} shows the maximum recall we can achieve with the algorithms presented in this paper.
Note that no algorithm achieve a coverage of 100\% as there is a small set of mentions with exactly the same span\footnote{This can be easily fixed by collpasing theses mentions, a standard trick used in the constituency parsing literature, see \cite{stern2017span}} and mentions that overlap partially.
We observe that the loss of coverage for our quadratic-time algorithm is negligible compared to the cubic-time algorithm for all datasets.

\textbf{Timing.}
We implemented the three algorithms in C++ and compare their running time for MAP inference in Figure~\ref{fig:time}.
The proposed algorithm is way faster than the CYK-like.
If we would parse only sentences of 300 words and we only consider the time spend in the decoding algorithm (i.e.\ ignoring the forward pass in the neural network), the CYK-like algorithm couldn't even decode 50 sentences in a second whereas our algorithm could decode more than 1500 sentences on an Intel Core i5 (2.4~GHz) processor.
As such, we hope that our algorithm will allow future work to consider NER on longer spans of text.

\textbf{Neural architecture and hyperparameters.}
Our neural network is composed of a finetuned \textsc{Bert} model\footnote{\emph{bert-base-uncased} as distributed at \url{https://huggingface.co/bert-base-uncased}} \cite{devlin2019bert} followed by 3 bidirectional \textsc{Lstm} layers \cite{lstm} with a hidden size of 400.
When the \textsc{Bert} tokenizer splits a word, we use the output embedding of the the first token.
Mention weights (i.e.\ values in vector $\vw$) are computed using two biaffine layers \cite{dozat2017biaffine}, one labeled and one unlabeled, with independent left and right projections of dimension 500 and \textsc{Relu} activation functions.

We use a negative log-liklihood loss (i.e.\ \textsc{Crf} loss) with $0.1$-label smoothing \cite{labelsmoothing}.
The learning rate is $1\times 10^{-5}$ for \textsc{Bert} parameters and $1\times10^{-3}$ for other parameters.
We use an exponential decay scheduler for learning rates (decay rate of $0.75$ every $5000$ steps).
We apply dropout with probability of $0.1$ at the output of \textsc{Bert}, $\textsc{Lstm}$ layers and projection layers.
We keep the parameters that obtains the best F1-measure on development data after 20 epochs.


\textbf{Results.}
We report experimental results in Table~\ref{tab:results}.
Note that our goal is not to establish a novel SOTA for the task but to assess whether our quadratic-time algorithm is well-suited for the nested NER problem, therefore we only compare our models with recent work using the same datasplit and comparable neural architectures (i.e.\ \textsc{Bert}-based and without lexicalization).
\textbf{Any method that modifies the cubic-time parser to improve results can be similarly introduced in our parser.}
Our implementation of the CYK-like cubic-time parser obtains results close to comparable work in the literature.
Importantly, we observe that, with the proposed quadratic-time algorithm, F1-measure results are (almost) the same on \textsc{Genia} and the the degradation is negligible on \textsc{Ace-2004} and \textsc{Ace-2005} (the F1-measure decreases by less than 0.5).

\section{Conclusion}

In this work, we proposed a novel quadratic-time parsing algorithm for nested NER, an asymptotic improvement of one order of magnitude over previously proposed span-based algorithms.
We showed that the novel search-space has a good coverage of English datasets for nested NER.
Despite having the same time-complexity than semi-Markov models, our approach achieves comparable experimental results to the cubic-time CYK-like algorithm.

As such, we hope that our algorithm will be used as a drop-in fast replacement for future work in nested NER, where the cubic-time algorithm has often been qualified of slow.
Future work could consider the extension to lexicalized mentions.

\section*{Limitations}

An obvious limitation of our work is the considered search space.
Although we showed that it is well suited for the data used in practice by the NLP community, this may not hold in more general settings.

Moreover, we only experiment in English.
We suspect that similar results would hold for morphologically-rich languages as we expect, in the latter case, that constituents are shorter (i.e.\ morphologically-rich languages heavily rely on morphological inflection, so we expect more mentions spanning a single word), see \cite[][Section 1.2 and Table 1.1]{haspelmath2013understanding}.
However, this is not guaranteed and future work needs to explore the multilingual setting.

Finally, in this work we do not consider discontinuous mentions, which is an important setting in real world scenario.

\section*{Acknowledgments}

We thank François Yvon, Songlin Yang and the anonymous reviewers for their comments and suggestions.
 This work benefited from computations done on the Saclay-IA platform and on the HPC resources of IDRIS under the allocation 2022-AD011013727 made by GENCI.
 
 \emph{I apologize to the reviewers for not adding the supplementary experimental results: being the sole author of this article, I unfortunately did not find the time to do so.}

\bibliography{refs}
\bibliographystyle{acl_natbib}

\end{document}